\documentclass[a4paper]{article}
\usepackage[T1]{fontenc}
\usepackage[english]{babel}
\usepackage[nocompress]{cite}
\usepackage[pdftex]{graphicx}
\graphicspath{figures/}
\DeclareGraphicsExtensions{.pdf,.jpeg,.png}
\usepackage{amsmath}
\usepackage{amssymb}
\usepackage{algorithmic}
\usepackage{array}
\usepackage[caption=false,font=footnotesize,labelfont=sf,textfont=sf]{subfig}
\usepackage{url}

\usepackage{authblk}
\usepackage[smaller,nolist]{acronym}
\usepackage{booktabs}
\usepackage{lipsum}
\usepackage{cleveref}

\begin{acronym}
  \acro{NPR}{Neighborhood Preserving Ratio}
  \acro{PCA}{Principal Component Analysis}
  \acro{SVD}{Singular Value Decomposition}
  \acro{MDS}{Multidimensional Scaling}
  \acro{UMAP}{Uniform Manifold Approximation and Projection}
  \acro{t-SNE}{t-distributed Stochastic Neighbor Embedding}
\end{acronym}{}

\newcommand{\RR}{\mathbb{R}}
\newcommand{\cX}{\mathcal{X}}
\newcommand{\cK}{\mathcal{K}}

\begin{document}

\title{An evaluation framework for dimensionality reduction through sectional curvature}

\author[1]{Raúl~Lara-Cabrera}
\author[2]{Ángel~González-Prieto}
\author[3]{Diego Pérez-López}
\author[3]{Diego Trujillo}
\author[1]{Fernando~Ortega}

\affil[1]{Departmento de Sistemas Informáticos, Universidad Politécnica de Madrid, Spain}
\affil[2]{Department of Algebra, Geometry and Topology, Universidad Complutense de Madrid, Spain, and the Instituto de Ciencias Matem\'aticas (CSIC-UAM-UCM-UC3M), Spain}
\affil[3]{Knowledge Discovery and Information Systems (KNODIS) Research Group, Universidad Politénica de Madrid, Spain}
\date{}                     
\setcounter{Maxaffil}{0}
\renewcommand\Affilfont{\itshape\small}

\maketitle

\begin{abstract}
Unsupervised machine learning lacks ground truth by definition. This poses a major difficulty when designing metrics to evaluate the performance of such algorithms. In sharp contrast with supervised learning, for which plenty of quality metrics have been studied in the literature, in the field of dimensionality reduction only a few over-simplistic metrics has been proposed. In this work, we aim to introduce the first highly non-trivial dimensionality reduction performance metric. This metric is based on the sectional curvature behaviour arising from Riemannian geometry. To test its feasibility, this metric has been used to evaluate the performance of the most commonly used dimension reduction algorithms in the state of the art. Furthermore, to make the evaluation of the algorithms robust and representative, using curvature properties of planar curves, a new parameterized problem instance generator has been constructed in the form of a function generator. Experimental results are consistent with what could be expected based on the design and characteristics of the evaluated algorithms and the features of the data instances used to feed the method.
\end{abstract}

\section{Introduction}\label{sec:introduction}

Unsupervised machine learning algorithms often suffer from a lack of performance metrics. This absence of metrics stems directly from the unavailability of ground truth labels for the training loop. In contrast, there are numerous options for assessing the quality of supervised learning models. Thanks to the ground truth labels, it is possible to make exhaustive comparisons between different models designed to solve a given problem, being possible not only to rank different algorithms, but also to quantify their performance.

With the consistent increase in the information to be processed, there is a growing interest in dimensionality reduction techniques. Not having a metric to evaluate the performance of these algorithms makes it very difficult to decide which of all of them is the most suitable for a given problem, beyond empirical experimentation.

Even in the domain of unsupervised learning, some metrics do exist. For clustering algorithms, there are metrics such as Silhouette Coefficient~\cite{Rousseeuw1987Nov}, Calinski-Harabasz Index~\cite{Calinski1974Jan}, and Davies-Bouldin index~\cite{Davies1979Apr}. However, for dimensionality reduction, to our best knowledge the only proposed metric is known as \ac{NPR}~\cite{4378396,li2019applying}.

\ac{NPR} is based on computing how many of the neighbors of a given point in our data are still neighbors after applying the dimensionality reduction for every point. This method, although somewhat effective, is highly inefficient due to the quadratic nature of determining the distance matrix required for neighborhood-based algorithms. Moreover, it over-simplifies the dimensionality reduction problem, since preserving the neighbourhood of a point provides no information about the global topology of the data. For instance, if a hyperplane in a high dimensional space is bent so that after the dimensionality reduction it self-intersects, then the \ac{NPR} will be high (close points are mapped to close points) but the global topology of the data has been completely destroyed.
Furthermore, the behaviour of \ac{NPR} is heavily limited by the selection of the neighborhood size hyperparameter.

To address this problem, in this paper we introduce a novel dimensionality reduction performance metric based on sectional curvature. This metric aims to serve as a benchmarking tool to objectively and quantitatively compare and measure dimensionality reduction algorithms.

\section{Problem instance generator}\label{sec:problem-generator}

The aim of the present work is to design a quality metric based on the concept of curvature to quantify the performance of dimension reduction techniques. To test its usefulness, this metric has been used to evaluate the performance of the most commonly used dimension reduction algorithms in the state of the art.

To make the evaluation of the algorithms robust and representative, a parameterized problem instance generator has been constructed in the form of a function generator named $\mathit{makegen}$. The generator $\mathit{makegen}$ returns a smooth function that is an immersion $\Phi: \RR^n \to \RR^m$ with $m > n$ whose image is an immersed sub-manifold $M_{\Phi} \subseteq \RR^m$ (the `underlying manifold' of the data). Recall that here immersion means that the differential of $\Phi$ has maximal rank. Hence, given a finite set of points $x_1, \ldots, x_N \in \mathbb{R}^n$ we obtain a collection of higher dimensional points $\Phi(x_1), \ldots, \Phi(x_N) \in \mathbb{R}^m$.

The $\mathit{makegen}$ generator allows us to `bend' the function $\Phi$ according to a collection of prescribed curvatures on each of the axis of $\RR^n$. In the implementation considered in this paper, the set of possible types of curvature is
$$
    \mathbb{K}=\left \{ \textrm{logistic}, \textrm{polyroll}, \textrm{sine}, \textrm{circle}, \textrm{flat} \right \}.
$$
Each possible type of curvature is configurable with a parameter, $\theta \in \RR$ so that the larger the value of $\theta$, the more curved is the given axis leading to harder synthetic datasets. Finally, to increase variability, rotation and translation, both random, are applied to the points forming the input data set. Like the magnitude of the curvature, the amount of random rotation and translation is also a global configurable real parameter $\eta$.

In this way, for fixed dimensions $n,m$ with $m > n$, the $\mathit{makegen}$ procedure is a function
$$
    \begin{array}{cccc}
        \mathit{makegen}: & \mathbb{K}^n \times \RR^n \times \RR & \to & C^\infty(\RR^n, \RR^m)\\
        & (\tau_i, \theta_i, \eta) & \mapsto & \Phi_{\tau_i, \theta_i, \eta}
    \end{array}\label{eq:makegen}
$$
where $ C^\infty(\RR^n, \RR^m)$ is the space of differentiable maps $\Phi: \RR^n \to \RR^m$ (in a more precise sense, $\mathit{makegen}$ returns a random variable that, when sampled, outputs an element of $C^\infty(\RR^n, \RR^m)$).

\subsection{Mathematical formulation of the generative model}\label{sec:math-gen}

The way the map $\Phi$ generated by the procedure $\mathit{makegen}$ is based on a standard result of reconstruction of curves. For the convenience of the reader, we review here the basic notions of the geometry of plane curves. For a more detailed account, please refer to \cite{do2016differential}.

A plane curve is a differentiable map $\gamma: I \to \mathbb{R}^2$, where $I \subseteq \mathbb{R}$ is a connected open interval. The curve is said to be regular if $\gamma'(t) \neq 0$ for all $t \in I$. In this case, $\gamma$ can be reparameterizedin terms of a new parameter $s = s(t)$ such that $||\gamma'(s)|| = 1$ for all $s$. Such a parametrization is called a parametrization by arc-length. Given a curve $\gamma(s)$ parameterized by arc-length, let $\mathbf{n}(s)$ be the normal vector at $\gamma(s)$ obtained by twisting $\gamma'(s)$ to the left $90$ degrees. The curvature $\kappa(s)$ is then defined as the scalar product $\kappa(s) = \mathbf{n}(s) \cdot \gamma''(s)$. In other words, $\kappa$ measures (with sign) the amount of variation of $\gamma(s)$ with respect to a straight line.

A key result is that this procedure also works in the other way around. Given a differentiable function $\kappa: I \to \mathbb{R}$, there exist a curve $\gamma_\kappa(s)$ parameterized by arc-length with curvature function $\kappa$. Explicitly, it is given by
$$
    \gamma_\kappa(s) = \left(\int_{a_1}^s \cos\left(\alpha_\kappa(t)\right)\, dt, \int_{a_2}^s \sin\left(\alpha_\kappa(t)\right)\,dt\right),
$$
where $\alpha_\kappa(t) = \int_{b}^t \kappa(u) \,du$. This curve is unique up to a rigid move of $\RR^2$, corresponding to fixing the integration initial points $a_1, a_2, b$.

In the proposed solution, we have fixed a collection of representative curvatures to be applied. They can be tuned through a parameter $\theta$ that sets up the growth of the curvature. The larger the value of $\theta$, the more prominent the growth. Explicitly, the different types of curvature that can be applied to generate the problem instances are the following:

\[
\begin{aligned}
    \textrm{logistic}^\theta(s)&=\frac{10 \theta}{1+e^{-0.5 s}},\nonumber\\
    \textrm{polyroll}^\theta(s)&=4\theta(s+1)^{2\theta},\nonumber\\
    \textrm{sine}^\theta(s)&=(5+10\cdot (\theta-1))\cdot \sin{(2\pi s)},\nonumber\\
    \textrm{circle}^\theta(s)&=2\pi \theta,\nonumber\\
    \textrm{flat}^\theta(s)&=0.\nonumber
\end{aligned}
\]

\subsection{Construction of the generative model}\label{sec:generative-model}

In this section, we shall explain how the ideas of~\cref{sec:math-gen} can be applied for constructing the problem instance generator $\mathit{makegen}$. Suppose that $\kappa_1, \ldots, \kappa_n: \RR \to \RR$ are curvature functions and let $\gamma_{\kappa_1}, \ldots, \gamma_{\kappa_n}$ be curves with this prescribed curvature, reconstructed as in~\cref{sec:math-gen}. These curves can be extended to spatial curves $\hat{\gamma}_{\kappa_i}: \mathbb{R} \to \RR^m$ by padding $\gamma_{\kappa_i}$ with $i-1$ zeroes on left and $m-i-1$ zeroes on the right.

In this manner, we can assemble them together into the function
$$
    \Phi_0(\kappa_1, \ldots, \kappa_n) = \sum_{i=1}^n \hat{\gamma}_{\kappa_i}: \RR^n \to \RR^m.
$$

However, notice that for $m > n+1$, the function $\Phi_0(\kappa_1, \ldots, \kappa_n)$ constantly vanishes on its last component. To overcome this problem, we randomly choose an orientation-preserving orthogonal transformation $R: \RR^m \to \RR^m$ (for instance, drawn following the Haar distribution on the special orthogonal group $\text{SO}(m)$ \cite{stewart1980efficient}). Finally, let $Z_{\eta}$ be a multivariate normal random variable with zero mean and diagonal covariance matrix $\eta\textrm{Id}$. 

Denote by $\kappa^\theta$ any of the types $\kappa \in \mathbb{K}=\left \{ \textrm{logistic}, \textrm{polyroll}, \textrm{sine}, \textrm{circle}, \textrm{flat} \right \}$ with tunning parameter $\theta$.
With these notions at hand, the $\mathit{makegen}$ procedure assigns
$$
    (\kappa_1, \ldots, \kappa_n, \theta_1, \ldots, \theta_n, \eta) \mapsto R \circ \Phi_0(\kappa_1^{\theta_1}, \ldots, \kappa_n^{\theta_n}) + Z_\eta.
$$
Observe that this is a random variable that, when sampled, returns a function $\Phi: \RR^n \to \RR^m$. This is the map used for immersing the original dataset of $\RR^n$ into the larger dimensional space $\RR^m$.

\section{Evaluation of dimensional reduction}\label{sec:evaluation}

This section provides the mathematical rationale behind using the notion of curvature as a quality measure for evaluating dimensional reduction algorithms.

\subsection{An overview of the notion of curvature}\label{ref:overview-curvature}

In this section, we will give a panoramic view of some of the main ideas that lead to the formulation of the metric for the evaluation of algorithms of dimensional reduction. This will be a very sketchy introduction with limited mathematical content so, for a more thorough description, please refer to Appendix \ref{ref:app:crash-course}.

In our setting, we will deal with an open set of the euclidean space $U \subseteq \RR^n$ (typically, $U$ will be the hypercube $U=(0,1)^n$. A \emph{Riemannian metric} $g$ is a smooth (differentiable) function $g: U \to \RR^{n^2}$ such that, for any $x \in U$, the evaluation $g(x) \in \RR^{n^2}$, seen as a $n \times n$-bilinear form, defines a scalar product $g(x): \RR^n \times \RR^n \to \RR$. Recall that this statement encompasses two conditions:
\begin{enumerate}
    \item $g(x)$ is a symmetric matrix.
    \item $g(x)$ is positive-define, meaning that $g(x)(v,v) = v^t g(x) v \geq 0$ for all $v \in \RR^n$ and $g(x)(v,v)=0$ if and only if $v = 0$.
\end{enumerate}

Given a Riemannian metric $g$, there is a unique way of assigning to it a \emph{sectional curvature tensor} $K^g$. It is an array $K^g = (K_{ij}^g)$ with $K_{ij}^g:U \to \mathbb{R}$ smooth functions, one for each possible unordered set $\left\{i,j\right\}$ where $1 \leq i,j\leq n$ with $i \neq j$. Therefore, $K^g$ is an array of dimension
$$
    \begin{pmatrix}n \\ 2\end{pmatrix} = \frac{n(n-1)}{2}.
$$

Intuitively, for $x \in U$ the coefficient $K_{ij}^g(x) \in \RR$ measures how the metric $g$ `bends' the space around $x$ in the $2$-dimensional $(i,j)$-space. A value of $K_{ij}^g(x)=0$ means that the space is flat in this direction, $K_{ij}^g(x) > 0$ means that the space in `inflated' in this direction as a sphere (elliptic curvature), and $K_{ij}^g < 0$ means that the space is `contracted' (hyperbolic curvature).

For instance, we can consider on $U$ the usual constant scalar product as a Riemannian metric, that is
$$
    g_{0}(x) = \begin{pmatrix}
    1 & 0 & \ldots & 0 \\
    0 & 1 & \ldots & 0 \\
    \vdots & & \ddots & \vdots \\
    0 & \hdots &  & 1 \\
    \end{pmatrix}
$$
for all $x \in U$. For this metric $g_{0}$ we obtain constant sectional curvature $K_{ij}^{g_0} \equiv 0$ for all $i,j$ (and, indeed, it is the only metric with this property up to isometry). Thus, for a general Riemannian metric $g$, the larger the function $|K_{ij}^g|$, the more curved the space and more different it is with respect to the usual euclidean space with metric $g_0$.

Additionally, Riemannian metrics have good contravariant functorial properties with respect to smooth functions. Suppose that we have two open sets $U,V \subseteq \RR^n$ and let $f: U \to V$ be a smooth function between them. Then, given a metric $g$ on $V$, it is possible to pull it back to get the so-called \emph{pullback metric} $f^*g$ on $U$.

In some sense, $f^*g$ captures how $f$ is deforming the metric $g$. For instance, the map $f: (U, g') \to (V, g)$ is an isometry if and only if $g' = f^*g$. Therefore, the sectional curvature of the pullback metric
$$
    K^{f^*g}_{ij}: U \to \RR
$$
measures the extend to which $f$ is deforming the metric $g$.

In particular, if we take $g = g_0$ to be the standard euclidean metric, then the pullback sectional curvature $K^{f^*{g_0}}_{ij}$ means how $f$ is bending the usual flat space: the larger the value of $|K^{f^*{g_0}}_{ij}|$, the more deformed is the space under $f$. Indeed, $|K^{f^*{g_0}}_{ij}| \equiv 0$ if and only if $f^*g_0 = g_0$ (up to rigid move), which means that $f$ is an isometry. Examples of isometries in $U$ are translations and rotations, the so-called rigid moves.

\subsection{Curvature as a metric for dimensional reduction}\label{sec:curvature-as-metric}

In our evaluation setting, suppose that we have a fixed problem instance generator
$$
    \Phi: U \subseteq \RR^n \to \RR^m
$$
for some $m > n$, say created with the method of~\cref{sec:problem-generator}. In the other direction, a dimensional reduction algorithm can be understood as a function
$$
    \Psi: \RR^m \to U \subseteq \RR^n
$$
that `projects' the points in the high dimensional space $\RR^m$ to the lower dimensional space $\RR^n$. Henceforth, we can compose both functions to get an endomorphism
$$
    f = \Psi \circ \Phi: U \to U.
$$

The key idea of the proposed evaluation framework is the following: the more similar the function $f$ is to an isometry, the better the dimensional reduction algorithm. Indeed, an ideal dimensional reduction algorithm should recover the original dataset up to rigid moves, namely, translation and rotation. This divergence of $f$ from being an isometry can be measured through the pullback sectional curvature introduced in~\cref{ref:overview-curvature} as the non-negative functions
$$
    |K_{ij}^{f^*g_0}|: U \to \RR.
$$
To summarize the size of these functions, we consider their $L^2$-norm, giving rise to the proposed quality measure
$$
    \cK(\Phi,\Psi) = \left[ \sum_{i \neq j} \left(\int_U \left(K_{ij}^{f^*g_0}\right)^2\,dx_1\ldots dx_n\right)\right]^{1/2}.
$$
Recall that this coefficient depends both on the chosen dataset generator $\Phi$ and on the dimensional reduction algorithm $\Psi$. We have that $\cK(\Phi,\Psi) = 0$ if and only if $f$ is an isometry, that is, if $\Psi$ preserves the dataset generated by $\Phi$ up to a rigid move.

\begin{figure}[t]
    \centering
    \includegraphics[width=0.6\columnwidth]{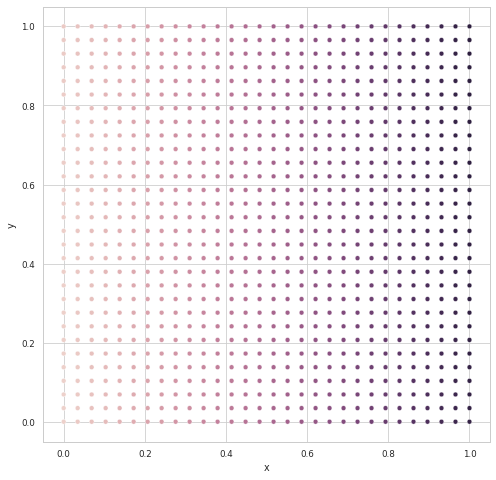}
    \caption{Grid of equispaced points in $\mathbb{R}^2$ that is the original data set of the metric.}
    \label{fig:grid}
\end{figure}

\subsection{Estimation of the Riemannian metric}

Despite the explicit description of the proposed quality measure in~\cref{sec:curvature-as-metric}, in practical applications, the dimensional reduction function $\Psi: \RR^m \to \RR^n$ is unknown. Indeed, standard dimensional reduction algorithms do not return a globally defined function $\Psi$ but instead they only compute the projection of a bunch of points.

This section addresses this problem and presents several methods to compute the quality measure $\cK(\Phi,\Psi)$ from the limited knowledge.

\begin{figure*}[t]
    \makebox[\textwidth][c]{\includegraphics[width=1.5\textwidth]{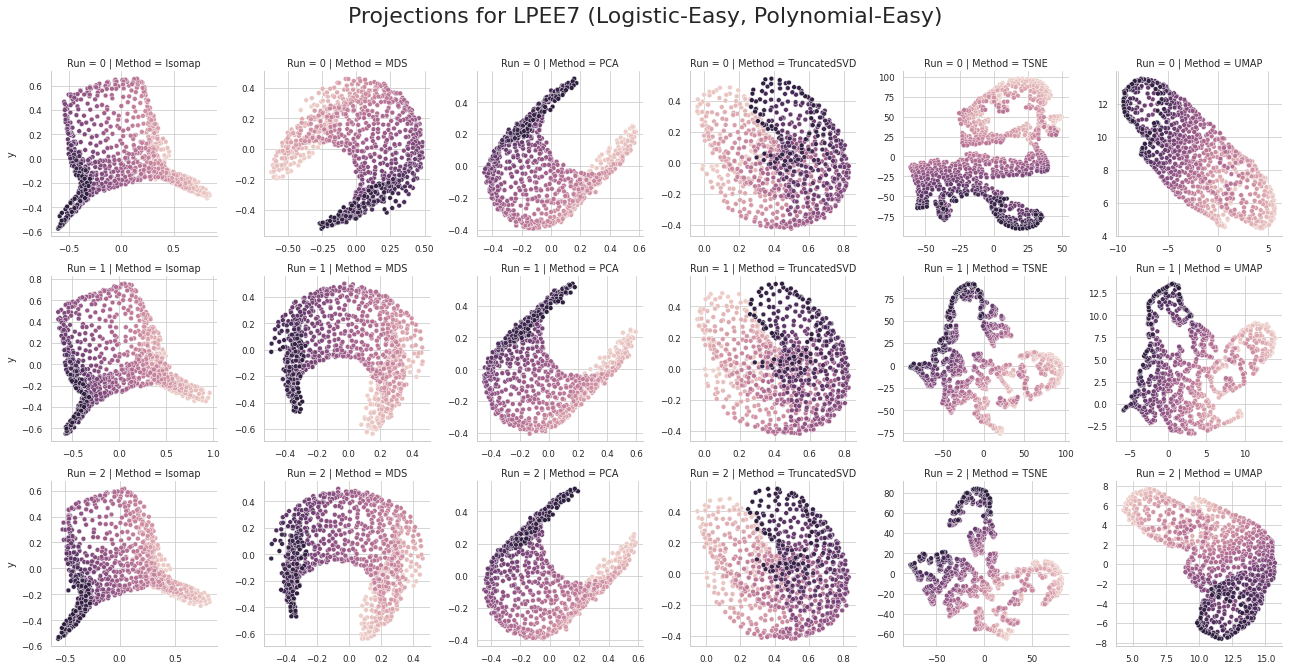}}%
    \caption{Projections of the LPEE7 instance in three independent runs by every algorithm.}
    \label{fig:projections}
\end{figure*}

\subsubsection{Interpolation of the function}\label{sec:interpolation-function}

The first way to address the aforementioned problem is to consider it as an interpolation problem: The function $f = \Psi \circ \Phi: U \to U$ is only known at finitely many points (the image of the generated points through the dataset generator) so we can straightforwardly apply any interpolation procedure to obtain a reasonable guess of $f$ on the unknown points.

However, there is a crucial point to be taken into account. As shown in Appendix \ref{ref:app:crash-course}, the computation of the sectional curvature requires to compute two derivatives of the Riemannian metric (one for obtaining the Christoffel symbols and another one for calculating the Riemann tensor). Moreover, the calculation of the pullback metric also involves derivatives of $f$ (Appendix \ref{sec:curvature}). Hence, it is not only necessary to accurately interpolate the value of $f$ but also of all its $n^3$ derivatives of order $3$. This may lead to instability problems due to the depth of the required interpolation.

The first solution we propose is to use an order $3$ spline interpolation~\cite{marsden1974cubic} to recover the smooth function $f$. The preliminary experiments show that the stability of the interpolation is satisfactory.

\subsubsection{Interpolation of the metric}

An alternative approach to the interpolation problem is to forget about the function $f = \Psi \circ \Phi$ and to focus on the interpolation of the pullback Riemannian metric $f^* g_0$. This has the advantage that the number of derivatives that need to be estimated for computing the curvature is lower, leading to smaller numerical errors. Indeed, the curvature tensor involves three derivatives of $f$, but only two derivatives of the metric $f^*g_0$. On the other hand, this estimation procedure will require to fix an extra hyperparameter $K$ which is a positive integer indicating the number of neighbors to be explored.

As explained in Appendix \ref{sec:pullback-metric}, the pullback metric $f^*g_0$ is given, for tangent vectors $u,v \in \RR^n$ at $x \in U$, by
$$
    f^*g_0(x)(u,v) = \left.\frac{\partial f}{\partial u}\right|_{x} \cdot \left.\frac{\partial f}{\partial v}\right|_{x},
$$
where $\frac{\partial f}{\partial u} = \left(\frac{\partial f_1}{\partial u}, \ldots, \frac{\partial f_n}{\partial u}\right)$ is the vector of the partial derivatives of $f$ in the direction $u$ and analogously for $v$, and $\cdot$ denotes the standard scalar product in $\RR^n$. The key idea of the method is to estimate the partial derivatives $ \frac{\partial f}{\partial u}$ using $K$ neighbors.

To be precise, fix $x \in U$ and let $x_1, x_2, \ldots, x_K \in U$ be the $K$ nearest points to $x$ among those whose value of $f(x_i) = \Psi(\Phi(x_i))$ is known (that is, the $K$ nearest generated points around $x$). These neighbors give rise to $K$ tangent vectors at $x$, namely $v_1 = x_1 - x, v_2 = x_2 - x, \ldots, v_K = x_K - x$. Up to order $2$, the derivative of $f$ in the direction $v_i$ can be thus estimated by
$$
    \left.\frac{\partial f}{\partial v_i}\right|_{x} \approx f(x_i) - f(x).
$$
Therefore, the value pullback Riemannian metric can be estimated at the $K^2$ pairs $(v_i, v_j)$ by
\begin{equation}\label{eq:approx-pullback-derivative}
    f^*g_0(x)(v_i,v_j) \approx \left(f(x_i) - f(x)\right) \cdot \left(f(x_j) - f(x)\right).
\end{equation}

On the other hand, the pullback metric $f^*g_0$ at $x$ is determined by a $n \times n$ matrix $A = (A_{ij})$ such that for all tangent vectors $u,v \in \RR^n$
\begin{equation}\label{eq:matrix-form-pullback}
    f^*g_0(x)(u,v) = u^tAv.
\end{equation}
Hence, putting together (\ref{eq:approx-pullback-derivative}) and (\ref{eq:matrix-form-pullback}), approximately we have that
$$
    v_i^tAv_j -  \left(f(x_i) - f(x)\right) \cdot \left(f(x_j) - f(x)\right) \approx 0.
$$
If $K > n$, this is an overdeterminated system of linear equations for $A$ which can be solved approximately by several numerical methods. In the present work, we chose least squares~\cite{levenberg1944method} to compute $A$. 

Working similarly at all the points $x \in U$ in the generated grid, we are able to estimate the pullback metric $f^*g_0$ on a grid of points. Now, applying any interpolation method (namely, spline interpolation), the first two derivatives of $f^*g_0$ at the grid points can be obtained as in Section \ref{sec:interpolation-function}.

\section{Experimental results}\label{sec:experiments}

To assess the usefulness of the curvature-based metric, this measure has been used to evaluate the performance of several dimensional reduction algorithms from the state of the art. To do so, we have constructed a set of problem instances as varied as possible, both in the types of curvature applied and in the difficulty of the problem instances. Starting from the assumption that our original data set is a grid of equispaced points in $\mathbb{R}^2$ (see~\cref{fig:grid}), we immerse them into $\mathbb{R}^7$ using a stochastic map $\Phi: \mathbb{R}^2 \to \mathbb{R}^7$. These maps have been exhaustively generated for all possible combinations of the following parameters:

\begin{itemize}
    \item Possible combinations of the curvature types $\mathbb{K}$ defined in~\cref{sec:problem-generator} are taken two by two, since the original data set has dimension $2$.
    \item Possible combinations taken two by two of the curvature difficulties corresponding to the parameter $\theta$ that sets up the growth of the curvature (see~\cref{sec:math-gen}). The values considered are easy ($\theta = 1.2$) and hard ($\theta = 1.8$).
    \item The standard deviation of $Z_{\eta}$ is $\eta = 0.01$ (see~\cref{sec:generative-model}).
\end{itemize}

This makes a total of 60 different problem instances.

\begin{figure}[t]
    \centering
    \includegraphics[width=\columnwidth]{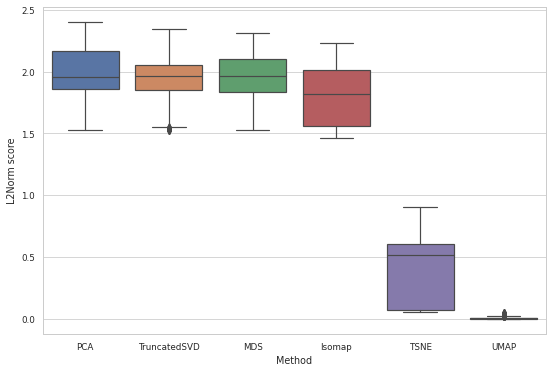}
    \caption{$L^2$-norm curvature score $\mathcal{K}$ of each dimensionality reduction method over all instances and independent runs carried out during the experiments. The lower, the better.}
    \label{fig:global_boxplot}
\end{figure}

Once the problem instances were obtained, each of them was used as input to the following dimension reduction algorithms:

\begin{itemize}
    \item \ac{PCA}~\cite{Wold1987Aug}.
    \item Truncated \ac{SVD}: a variant of \ac{SVD} that only computes the $k$ largest singular values, where $k$ is a user-specified parameter.
    \item Metric \ac{MDS}~\cite{Kruskal1964Jun}.
    \item ISOMAP~\cite{Tenenbaum2000Dec}.
    \item \ac{t-SNE}~\cite{vanderMaaten2008}.
    \item \ac{UMAP}~\cite{McInnes2018Feb}.
\end{itemize}

\begin{figure}[t]
    \makebox[\textwidth][c]{\includegraphics[width=1.5\textwidth]{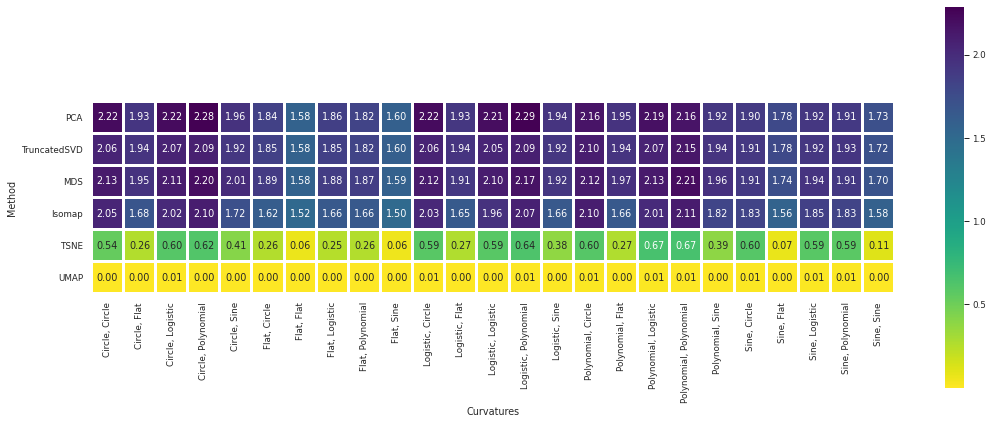}}%
    \caption{Median $L^2$-norm curvature score $\mathcal{K}$ of each dimensionality reduction method over every complexity instance type and independent runs carried out during the experiments. The lower, the better.}
    \label{fig:heatmap}
\end{figure}

The process of evaluating the performance of a dimension reduction algorithm applied to a particular instance is described in the following steps:

\begin{enumerate}
    \item The problem instance (function obtained from $\mathit{makegen}$) is applied on the grid of equispaced points in $\mathbb{R}^2$, obtaining a dataset in $\mathbb{R}^7$.
    \item In order to compare the different dimensionality reduction methods in the fairest possible way, an optimization of the hyperparameters has been performed for those dimension reduction algorithms that have them (ISOMAP, \ac{t-SNE} and \ac{UMAP}). Optimization has been conducted by using the Nevergrad library~\cite{nevergrad}.
    \item The (best) dimensionality reduction algorithm is applied on the data set in $\mathbb{R}^7$ obtained in the first step to obtain a new data set in $\mathbb{R}^2$ (some examples in~\cref{fig:projections}).
    \item Using our quality metric $\mathcal{K}$ described in~\cref{sec:evaluation}, the curvature of the initial grid is compared to the data set resulting from applying the dimensionality reduction algorithm.
\end{enumerate}

Once the experiments have been carried out, it can be seen that the proposed metric allows us to establish a ranking of the different dimensionality reduction methods according to their performance. As can be seen in~\cref{fig:global_boxplot}, the \ac{UMAP} method together with t-SNE are the ones that obtain the best score, since they are able to preserve the curvature of the initial data set after its projection from a higher dimensional space. In contrast, the rest of the algorithms score orders of magnitude worse, due to their inability to preserve the original curvature of the data set.

Focusing on the capacity of the dimensional reduction algorithms depending on the complexity of the instance (see~\cref{fig:heatmap}), we can observe that \ac{UMAP} obtains good results regardless of the type of curvature applied to the dimensions. In the case of \ac{t-SNE}, although it reaches low values compared to the rest of the methods except \ac{UMAP}, it can be observed that in those instances with complex curvatures, such as circle, logistic and polynomial, the method presents more difficulties in preserving the curvature, as can be seen by the higher values of the metric. On the other hand, it is able to achieve good results in instances with easy (Sine) or no curvature (Flat). With respect to the rest of the methods analyzed, they obtain their best results with those instances in which no curvature has been applied to any of their dimensions, i.e. one of the curvatures applied is Flat. This result is consistent with the fact that \ac{PCA}, Truncated \ac{SVD} and Metric \ac{MDS} are quasi-linear models, so they should perform better when there is linearity in the manifold whose dimensionality is to be reduced. It is also compatible with the fact that ISOMAP is, in some sense, a simpler version of the \ac{t-SNE} algorithm, leading to the same pattern of results but one order of magnitude worse.

It is worth mentioning that, compared to the previous \ac{NPR} quality measure, our metric presents a much more robust behaviour and captures highly non-linear relations. At the end of the day, \ac{NPR} is only measuring to which extent is the projection algorithm continuous, in the sense that it only focuses on the fact that close points are mapped together. However, continuity is a much simpler concept than preserving the underlying geometry, as is required for good dimensionality reduction methods. For instance, mere continuity does not characterize whether the actual shape of figures in the data is preserved after projection. In sharp contrast, measuring metric properties does allow us to detect this potential distortion, and to punish those dimensionality reduction algorithms that produce it.

\section{Conclusion}
In this paper we present an evaluation metric for dimensional reduction algorithms based on the concept of curvature. Starting from a grid of equispaced points in $\mathbb{R}^2$ with constant and known curvature, we embed them into a submanifold of $\mathbb{R}^7$ made of parameterizable curvatures that take the data to a higher dimensionality. Subsequently, the algorithm to be evaluated is asked to reduce the dimensionality of the data set again to dimension $2$. The performance of the algorithm will be measured as the difference between the curvature of the original grid and the curvature of the dimensional reduction.

To test the usefulness of the metric we have generated a set of test instances combining different types of curvature as well as their modulation to cover as much variability as possible. We have then evaluated six state-of-the-art dimensionality reduction algorithms using the proposed metric, obtaining results consistent with the common consensus of the community about which ones are the most flexible methods. 
The authors of this work are committed to reproducible science. Both the problem instance generator and the metric have been published as a Python package\footnote{\url{https://pypi.org/project/manifoldy/}}. The reader can find the source code in a Github repository\footnote{\url{https://github.com/KNODIS-Research-Group/manifoldy}}.

This curvature-based quality measure provides, for the very first time, an objective way of characterising the quality of a dimensionality reduction algorithm, which we expect will be very useful to boost future research in this direction and to discriminate whether a new proposed method surpasses the baseline or not. Other pre-existing methods, such as \ac{NPR}, are not able to analyze such geometric features and stick to very simple continuity properties.
To provide a visual simile, suppose that our data represent the stars in the celestial vault and we want to project them into a plane. The \ac{NPR} metric would only analyze whether the stars of each constellation are mapped together, but it does detect if princess Andromeda is turned into the goat Capricornius. On the contrary, since our curvature measure is deeply based on metric information, it will penalize those algorithms that distort the constellations. In this sense, the curvature-based quality metric is able to recognize those methods able to provide a faithful celestial chart that converts Aquila into an eagle, Cancer into a crab, Libra into a balance and, in general, that capture the true shape of the constellations in the firmament in a much more geometric sense.

\section*{Acknowledgments}

This work was partially supported by the \textit{Comunidad de Madrid} under \textit{Convenio Plurianual} with the Universidad Politécnica de Madrid in the actuation line of \textit{Programa de Excelencia para el Profesorado Universitario} and \textit{Ministerio de Ciencia e Innovación} of Spain under the project PID2019-106493RB-I00 (DL-CEMG). The second author has been partially supported by the Madrid Government (Comunidad de Madrid – Spain) under the Multiannual
Agreement with the Universidad Complutense de Madrid in the line Research Incentive for
Young PhDs, in the context of the V PRICIT (Regional Programme of Research and Technological Innovation) through the project PR27/21-029 and by the Ministerio de Ciencia e Innovaci\'on Project PID2021-124440NB-I00 (Spain).

\newpage


\bibliographystyle{ieeetr}
\bibliography{preprint}

\appendix

\section{A crash course in Riemannian geometry}\label{ref:app:crash-course}

In this section, we will briefly review some concepts of Riemannian and differential geometry that will be needed for the formulation of the metric for the evaluation of algorithms of dimensional reduction. For simplicity, we restrict our attention to the framework that we will need for applications, in which the Riemannian manifolds are open subsets of the Euclidean space $U \subseteq \RR^n$. This leads to several important simplifications derived from the fact that the tangent bundle of the Euclidean space is trivial. The general case can be addressed similarly by seeing $U$ as a chart of an abstract manifold. For a more complete introduction, please refer to \cite{carmo1992riemannian,kobayashi1963foundations}.

\subsection{Riemannian metrics and connections}

Fix an open subset $U \subseteq \RR^n$. Recall from Section \ref{ref:overview-curvature} that a Riemannian metric $g$ is a smooth function $g: U \to \RR^{n^2}$ such that, for any $x \in U$, the evaluation $g(x) \in \RR^{n \times n}$ defines a scalar product $g(x): \RR^n \times \RR^n \to \RR$ in the sense that:
\begin{itemize}
    \item $g(x)$ is a symmetric matrix.
    \item $g(x)$ is positive-define, meaning that $g(x)(v,v) = v^t g(x) v \geq 0$ for all $v \in \RR^n$ and $g(x)(v,v)=0$ if and only if $v = 0$.
\end{itemize}
It is customary to write down explicitly the entries of the matrix of the metric as $g = (g_{ij})_{i,j=1}^n$, where $g_{ij} \in C^\infty(U)$ are smooth functions.

A Riemannian metric can be fed with a vector field to obtain functions. Recall that a vector field $X$ in $U$ is a smooth function $X: U \to \RR^n$, and the set of vector fields on $U$ will be denoted by $\cX(U)$. Then, given $X, Y \in \cX(U)$, we get a function $g(X,Y): U \to \RR$ given by $x \mapsto g(x)(X_x, Y_x)$.

A related concept is the one of a connection. A connection $\nabla$ is a map
$$
    \nabla: \cX(U) \times \cX(U) \to \cX(U),
$$
denoted by $(X,Y) \mapsto \nabla_XY$, such that for all $X, Y, Z \in U$ and function $f \in C^\infty(U)$, we have
\begin{itemize}
    \item (Linearity) $\nabla_{X + fY}Z = \nabla_XZ + f \nabla_YZ$.
    \item (Leibniz rule) $\nabla_{X}(Y + fZ) = \nabla_XY + \frac{\partial }{\partial Y}(f) Z + f\nabla_YZ$, where $\frac{\partial}{\partial Y} (f)= \textrm{grad}(f) \cdot Y$ denotes the partial derivative of $f$ in the direction $Y$.
\end{itemize}
A connection can be easily understood in terms of its local form. Let $e_1, \ldots, e_n \in \RR^n$ be the standard coordinate basis of $\RR^n$. Write the vector fields $X$ and $Y$ as
$$
    X = \sum_{i=1}^n \alpha_i e_i,\quad X = \sum_{j=1}^n \beta_j e_j,
$$
where $\alpha_i, \beta_j \in C^\infty(U)$ are smooth functions. Then, using the properties of the connection, we can write $\nabla_XY$ as
$$
    \nabla_XY = \sum_{i,j=1}^n \alpha_i \left(\frac{\partial \beta_j}{\partial x_i} + \sum_{k=1}^n \Gamma_{ij}^k e_k\right).
$$
Here, $\Gamma_{ij}^k$ is a collection of $n^3$ smooth functions that satisfy $\nabla_{e_i}e_j = \sum_{k=1}^n\Gamma_{ij}^k e_k$, known as the Christoffel symbols. Notice that the Christoffel symbols completely determine the connection.

The key interplay between connections and Riemannian metrics is that, given a Riemannian metric $g$, there is a unique connection $\nabla$ compatible with $g$, the so-called Levi-Civita connection. Explicitly, the compatibility means that that it satisfies, for all vector fields $X,Y, Z \in \cX(U)$:
\begin{itemize}
    \item (Compatibility with $g$) $\frac{\partial}{\partial X}g(Y,Z) = g(\nabla_XY, Z) + g(Y, \nabla_XZ)$.
    \item (Torsion-free) $\nabla_X Y - \nabla_Y X = [X,Y]$ where $[X,Y]$ is the Lie bracket vector field given by $[X,Y] = XY - YX$.
\end{itemize}
The Christoffel symbols $\Gamma_{ij}^k$ of the Levi-Civita connection are fully determined by the above-mentioned conditions. It can be easily checked that they are explicitly given by  
$$
\Gamma^k_{ij} =\frac{1}{2}\, \sum_{m=1}^n g^{mk} \left(
        \frac{\partial}{\partial x_j} g_{mi}
        +\frac{\partial}{\partial x_i} g_{mj}
        -\frac{\partial}{\partial x_m} g_{ij}
        \right),
$$
where $g^{mk}$ are the entries of the inverse matrix of the Riemannian metric $g=(g_{ij})$.

\subsection{Pulling-back metrics}\label{sec:pullback-metric}

Riemannian metrics have contravariant functoriality with respect to smooth maps. Let $U,V \subseteq \RR^n$ be open sets and let $f: U \to V$ be a smooth function between them. Given a Riemannian metric $g$ on $V$, we can get a metric $f^*g$ on $U$, called the \emph{pullback metric}. It is given as follows.

Let $Df: \RR^n \to \RR^n$ be the differential of $f$ that is, the matrix of partial derivatives
$$
    Df = \begin{pmatrix}
    \frac{\partial f_1}{\partial x_1} & \frac{\partial f_1}{\partial x_2} & \hdots & \frac{\partial f_1}{\partial x_n} \\
    \frac{\partial f_2}{\partial x_1} & \ddots & \hdots & \frac{\partial f_2}{\partial x_n} \\
    \vdots &  &  & \vdots \\
    \frac{\partial f_n}{\partial x_1} & \hdots & & \frac{\partial f_n}{\partial x_n} \\
    \end{pmatrix}
$$
where $f$ is written in coordinates as $f = (f_1, f_2, \ldots, f_n)$.

Using this map, given tangent vectors $v,w \in \RR^n$ to $U$ at $x \in U$, we can transfer them to tangent vectors $Df(v), Df(w) \in \RR^n$ to $U$ at $f(x)$. Hence, we define the pullback metric to be
$$
    (f^*g)(x)(v,w) = g(f(x))\left(Df(v), Df(w)\right)
$$

In the particular case that $g = g_0$ is the standard constant metric, we can easily observe that the components of $f^*g_0$ are
$$
    (f^*g_0)_{ij} = \nabla f_i \cdot \nabla f_j,
$$
where $\nabla f_i$ is the usual gradient of $f_i: U \to \RR$ and the dot denotes the usual scalar product.

\subsection{Curvature}\label{sec:curvature}

Given a connection $\nabla$, we can associate to it the so-called Riemann curvature tensor $R$, given for $X, Y, Z \in \cX(U)$ by
$$
    R(X,Y)Z = \nabla_Y\nabla_XZ - \nabla_X\nabla_YZ - \nabla_{[X,Y]}Z.
$$
Explicitly, in terms of the coordinates $e_1, \ldots, e_n$, if we consider vector fields $X=\sum_i \alpha_i e_i$, $Y=\sum_j \beta_j e_j$ and $Z=\sum_k \alpha_k e_k$ we have that
$$
    R(X,Y)Z = \sum_{i,j,k,l=1}^n R_{ijk}^l e_l,
$$
for some smooth functions $R_{ijk}^l \in C^\infty(U)$ called the components of $R$. They can be explicitly computed from the Christoffel symbols of the connection as 
$$
    R_{ijk}^l=\frac{\partial\Gamma_{ik}^l}{\partial x_j}-\frac{\partial\Gamma_{jk}^l}{\partial x_i}+\sum_{p=1}^n \big(\Gamma_{ik}^p\Gamma_{jp}^l-\Gamma_{jk}^p\Gamma_{ip}^l\big)
$$
From this Riemann curvature tensor, we can define the so-called sectional curvature. Let $X$ and $Y$ be vector fields that are linearly independent for all $x \in U$. The sectional curvature $K(X,Y)$ of the distribution of planes generated by $X$ and $Y$ is the smooth function on $U$ given by
$$
    K(X,Y) = \frac{g(R(X,Y)Y, X)}{\sqrt{g(X,X)g(Y,Y)-g(X,Y)^2}}.
$$
As always, we can write down $K$ in terms of coordinates as a bunch of functions $K = (K_{ij})$ where
$$
    K_{ij} = \frac{g(R(e_i,e_j)e_j, X)}{\sqrt{g(e_i,e_i)g(e_j,e_j)-g(e_i,e_j)^2}}.
$$
This quantity is computed for each pair of coordinated vectors $e_i, e_j$ with $i \neq j$, so these are $\begin{pmatrix}n \\ 2\end{pmatrix} = \frac{n(n-1)}{2}$ different functions.

\end{document}